This article has been accepted for publication in IEEE Robotics and Automation Letters. This is the author's version which has not been fully edited and content may change prior to final publication. Citation information: DOI 10.1109/LRA.2024.3376953

# Single-motor robotic gripper with multi-surface fingers for variable grasping configurations

Toshihiro Nishimura[1], *Member, IEEE*, Yosuke Suzuki[1], *Member, IEEE*,
Tokuo Tsuji[1], *Member, IEEE*, and Tetsuyou Watanabe[1], *Member, IEEE*

***Abstract*—This study proposes a novel robotic gripper with variable grasping configurations for grasping various objects. The fingers of the developed gripper incorporate multiple different surfaces. The gripper possesses the function of altering the finger surfaces facing a target object by rotating the fingers in its longitudinal direction. In the proposed design equipped with two fingers, the two fingers incorporate three and four surfaces, respectively, resulting in the nine available grasping configurations by the combination of these finger surfaces. The developed gripper is equipped with the functions of opening/closing its fingers for grasping and rotating its fingers to alter the grasping configuration— all achieved with a single motor. To enable the two motions using a single motor, this study introduces a self-motion switching mechanism utilizing magnets. This mechanism automatically transitions between gripper motions based on the direction of the motor rotation when the gripper is fully opened. In this state, rotating the motor towards closing initiates the finger closing action, while further opening the fingers from the fully opened state activates the finger rotation. This paper presents the gripper design, the mechanics of the self-motion switching mechanism, the control method, and the grasping configuration selection strategy. The performance of the gripper is experimentally demonstrated.**

## I. Introduction

THIS study presents a novel robotic parallel-jaw gripper that can alter finger surfaces and construct multiple grasping configurations using a single motor. A robotic gripper is an important end-effector in robotic manipulators, facilitating object handling [1]. Among robotic grippers, the two-fingered parallel-jaw configuration has emerged as the first choice, characterized by the simplicity of its control: simple closing and opening of the fingers. Another feature of most parallel grippers is that they are driven using a single actuator. A reduction in the number of actuators not only simplifies the control of the gripper, but also fosters a lightweight and compact gripper. Some commercial robotic grippers feature interchangeable finger parts [2] that allow users to flexibly customize their fingers for their specific target objects. A finger design specialized for the target objects enables the gripper to secure a stable grasp through a preferred grasping configuration. However, these specialized fingers lack the versatility to grasp objects beyond the pre-targeted ones. An alternative solution involves employing a tool changer [3][4] capable of swapping fingers for each target object. However, this approach requires the implementation of jigs and finger exchange operations, resulting in a larger entire system and increased takt time. To overcome this issue, this study proposes a new robotic gripper that incorporates functionality akin to that of the tool changer. Fig. 1 shows the developed gripper. The gripper features multiple surface fingers, with each finger possessing three or four distinct surfaces, each uniquely shaped. The fingers perform not only the conventional opening and closing functions, but also possess the capability for longitudinal rotation (Fig. 1(a)). This rotational action alters the

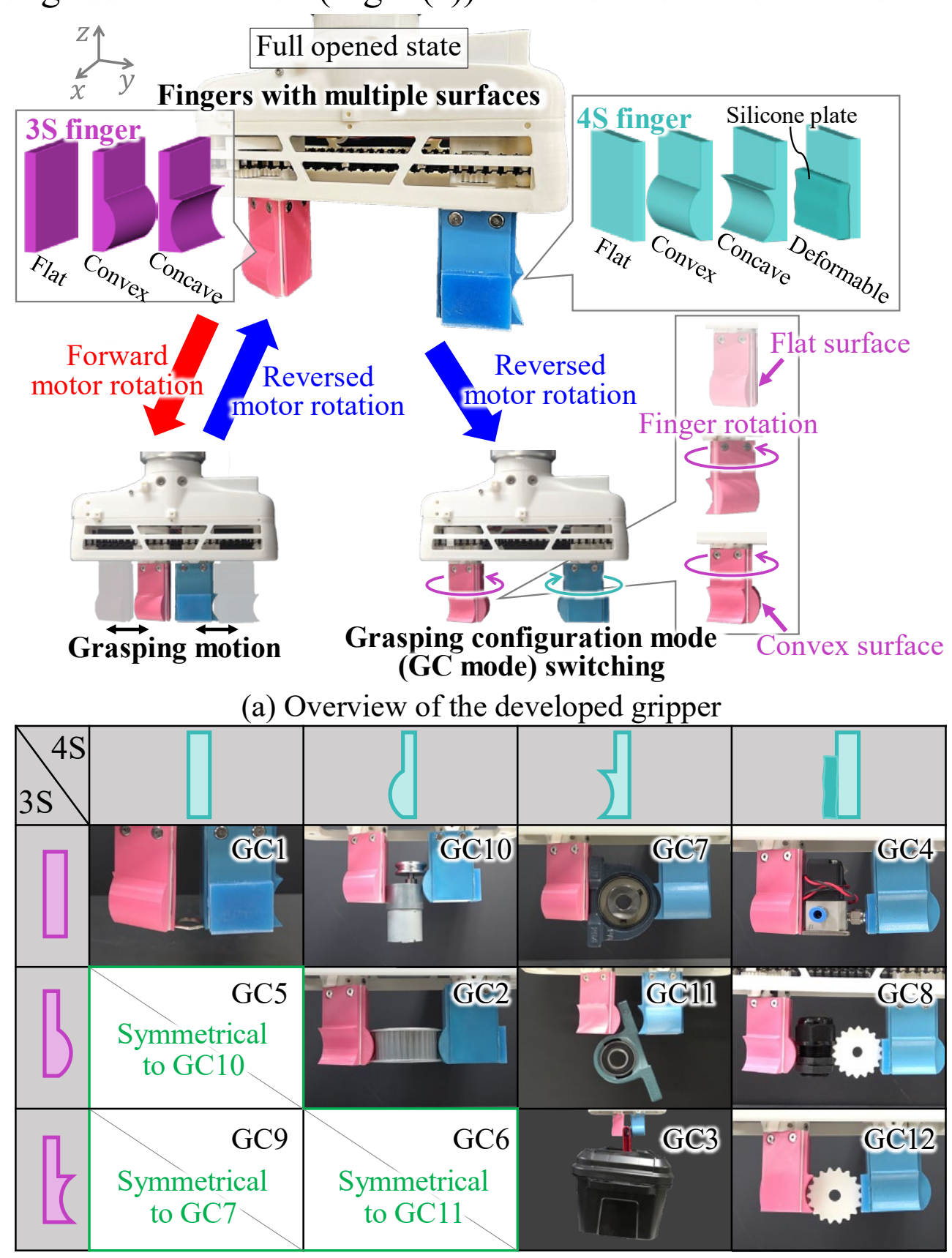

(a) Overview of the developed gripper

(b) Representative results of the grasping tests conducted while changing grasping configuration (GC) modes. Here, "GC*" refers to the GC mode number.

**Fig. 1.** Developed gripper.

Manuscript received: December 12, 2023; Revised: February 13, 2024; Accepted: March 6, 2024. Footnote sentence 2: This paper was recommended for publication by Editor Júlia Borràs Sol upon evaluation of the Associate Editor and Reviewers' comments. This work was supported by JSPS KAKENHI Grant Number JP22K14220. (*Corresponding authors: Toshihiro Nishimura; Tetsuyou Watanabe*)

[1]The authors are with the Faculty of Frontier Engineering, Institute of Science and Engineering, Kanazawa University, Kakuma-machi, Kanazawa city, Ishikawa 9201192, Japan (e-mail: tnishimura@se.kanazawa-u.ac.jp and te-watanabe@ieee.org)

finger surfaces, thereby changing the grasping configuration mode (GC mode), as shown in Fig. 1(b). The type of activated GC mode is determined by the combination of surfaces on each finger. Considering the symmetrical arrangement within the combination, nine GC modes were available in the proposed design. The challenge addressed in this study lies in achieving both closing/opening motions for grasping and rotational motion to change the GC mode using a single motor. This challenge enhances the versatility of robotic grippers while minimizing the overall robotic gripper system, including the weight, size, and wiring. To realize the two different motions using a single motor, a new self-motion-switching mechanism was developed. This mechanism automatically switches between these two motions. Basically, the opening/closing motion of the finger is generated by forward and reverse motor rotations, same as conventional parallel grippers. The motion switching is activated when the finger is fully open. When the motor turns in the direction of opening the fingers beyond the fully opened state, the self-motion switching mechanism is activated, leading to the alternation of the finger surface via the rotational motion of the fingers (Fig. 1(a)). The magnets in the switching mechanism ensure highly durable discrete motion switching.

Several studies have developed robotic grippers with variable grasping configuration for grasping various objects [1]. There are two main approaches for changing the grasping configuration: passive and active. The former method utilizes contact with objects or the environment such as a table. The introduction of flexible structures to the joints and links allows the finger posture to adapt to the shape of the target object. This self-adaptability can be regarded as a form of grasping configuration change. Tamamoto et al. proposed a gripper with self-adaptability using a differential gear [5]. In [6], self-adaptive grasping was realized using a linkage mechanism. The grippers developed in [7] and [8] exhibit the capability to conform their shape to that of an object upon contact. Our research group has also developed variable-grasping-mode grippers using contact with the environment [9][10]. As reported in these studies, the passive function of changing the grasping configuration was accomplished through a 1-DOF actuation. However, the extent of the available grasping configurations is restricted to changes in finger posture. In addition, these passive-type grippers use deformable elements such as springs and rubbers, which induce uncertainties in the positions and postures of the grasped objects. In the latter method, a change in the grasping configuration is achieved by employing a substantial number of actuators, as represented by a multi-fingered hand designed to resemble the human hand [11]. Various grippers that can actively transition into a predetermined set of grasping configurations have been developed. Jain et al. developed a soft gripper featuring retractable nails designed to achieve both power- and pinching-grasping functionalities [12]. Elangovan et al. proposed a multilink gripper with adaptive and parallel-jaw gripper modes [13]. In [14], a pneumatic gripper capable of envelope and suction grasping was developed. In [12][13][14], two or more actuators were employed to execute both grasping motion and mode changes. Although these grippers can change various grasping modes, their drawbacks lie in the enlargement of the gripper design and complexity of the control associated with an increase in the number of actuators. A 1-DOF gripper capable of actively changing grasping modes was developed [15]. However, only two grasping modes were achieved in [15]. Consequently, no attempt has been made to develop a robotic gripper capable of changing its grasping configuration in numerous ways with only a single motor.

Several self-motion switching mechanisms have been developed and implemented in robotic grippers. Ko developed a robotic gripper with a pull-in function activated by the contact of a finger with an object [16]. In [17], a gripper with a function similar to that in [16] was developed. The mechanism developed in [16] used the extension of an installed spring to generate a pull-in function; however, the input motor torque continued to be distributed to the grasping motion even after the pull-in function was activated. As the motor torque for the pull-in function increased, the grasping force was amplified, thereby increasing the load on the grasped object. Consequently, the spring-based mechanism fails to achieve discrete motion switching. In contrast, the approach presented in [17] utilizes frictional resistance to transition from grasping motion to pull-in motion. Frictional fixation generates two distinct states, slipped and unslipped, enabling discrete and differentiated motions. Similarly, our research group developed multi-functional grippers using a self-motion switching mechanism based on frictional resistance [18][19]. However, friction-based mechanisms face the challenge of reduced durability due to wear. Therefore, the development of a mechanism capable of achieving discrete motion switching with high durability remains an unresolved challenge.

## II. Robotic Gripper Design

### A. Functional requirement

The functional requirements of the proposed robotic gripper are as follows: 1) multiple grasping configuration changes and opening/closing actions using a singular motor; 2) accommodation of graspable objects ranging from heavy (>5 kg) to thin (<1 mm); and 3) a motion-switching mechanism devoid of wear-related concerns.

### B. Gripper design

Fig. 2 shows a three-dimensional computer-aided design (3D-CAD) model. A method was adopted in which the grasping configuration was changed by altering the finger surface. Thus, the fingers must translate for the opening and closing motions of the gripper, and rotate around the longitudinal axis for GC mode switching. As shown in Fig. 2(a), the developed gripper includes two finger units, a motor (black part in Fig. 2(a)), a roller chain (black), and a gripper body (gray). These two fingers are referred to as the 3S and 4S fingers, which possess three and four surfaces, respectively. The proposed design separates the finger bodies from the sprocket, enabling the single-motor drive to rotate the two fingers at different angles.

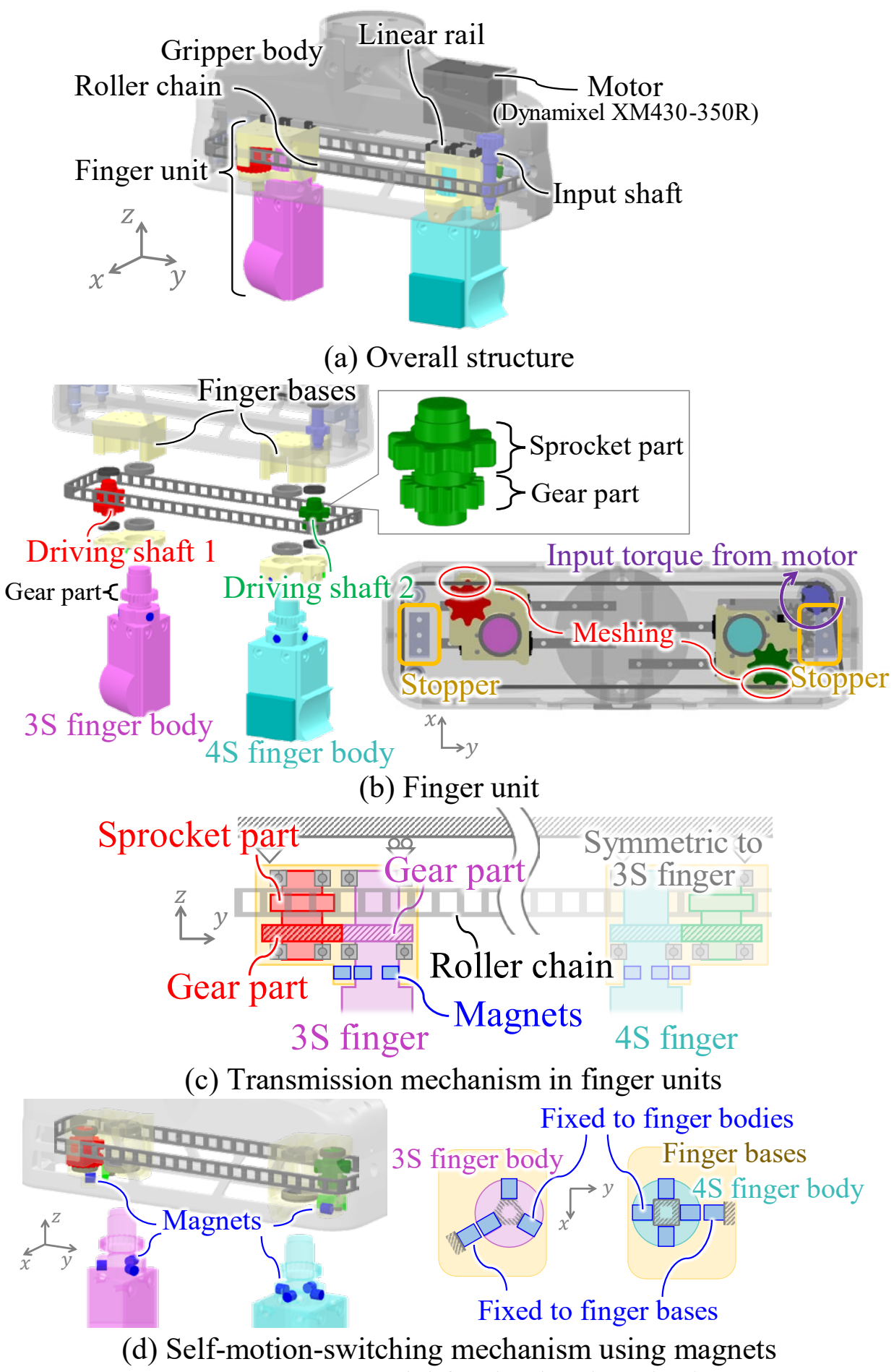

**Fig. 2.** 3D-CAD model for the developed gripper

This increases the number of the available GC modes. The 3S finger incorporates flat, convex, and concave-shaped surfaces made of rigid PLA resin. The 4S finger incorporates a deformable flat surface made of silicone (Dragon Skin 30) in addition to the aforementioned three-shaped surfaces. The finger units are attached to the gripper body via linear rails, enabling the translation of the finger units for opening and closing motions. The motor with an input shaft (purple) is mounted on the gripper body. The input shaft incorporates a sprocket part that meshes with the roller chain. As shown in Fig. 2(b), each finger unit comprises a finger body (pink and light blue), finger base (light yellow), driving shaft (red and green), and magnets (blue; see Fig. 2(d)). The driving shafts incorporate a sprocket part that engages the roller chain, allowing the transmission of motor torque to the finger units through the roller chain. The finger bodies also contain gear parts that mesh with the gear parts of the driving shafts. The transmission mechanism is shown in Fig. 2(c). The gear ratio between the gears is tuned such that the two fingers with different numbers of the surfaces face each other simultaneously. Magnets are installed to switch between the translation and rotation of the finger bodies (Fig. 2(d)). Three and four magnets are arranged in circular symmetry within the 3S and 4S finger bodies, respectively. In addition, a single magnet is incorporated into each finger base to establish an attractive force between the magnets on the finger body and those on the finger base. The position of each finger surface is aligned with the location of the magnet to ensure that any alteration in the position of the magnet induces a transition in the finger surfaces. Unless the torque applied to the finger body from the driving shafts exceeds the generated magnetic attraction, the finger unit performs the translational motion, i.e., opening and closing movements. If the torque applied to the finger body exceeds the magnetic attraction generated, the torque induces the rotation of the finger bodies, leading to switching of the finger surfaces, i.e., GC mode switching. The stoppers are positioned on the gripper body to limit the translational motion range of the finger unit during the grasping movements, as shown in Fig. 2(b). Continued motor rotation after the finger units/bases contact the stopper increases the motor torque and the torque applied to the finger bodies. This increased torque on the finger body triggers the activation of GC mode switching. Note that the GC mode switching could be triggered in situations where a significant grasping force is exerted while grasping an object. To prevent undesired GC switching, a ratchet mechanism is integrated into both the finger bodies and bases, as illustrated in Fig. 3. Each finger body incorporates a ratchet gear part, and each finger base is equipped with a pawl. This ratchet mechanism permits the rotation of the finger bodies only in the direction corresponding to the motor rotation when the fingers are open. Consequently, GC mode switching remains inactive when the finger units are closed to grasp the objects. The number of teeth on each ratchet gear corresponds to the number of surfaces on each finger, i.e., three and four. The engagement of the ratchet gear and pawl occurs when the finger surface is oriented towards the object, ensuring stable positioning of finger rotation.

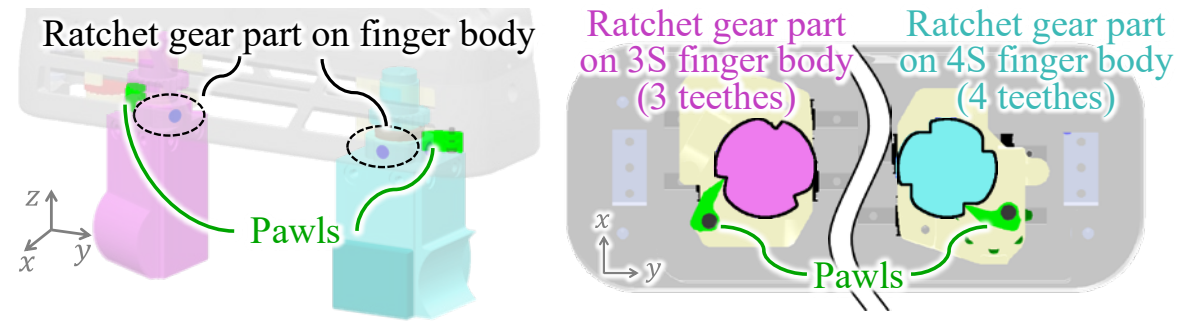

**Fig. 3.** Ratchet mechanism in the finger unit

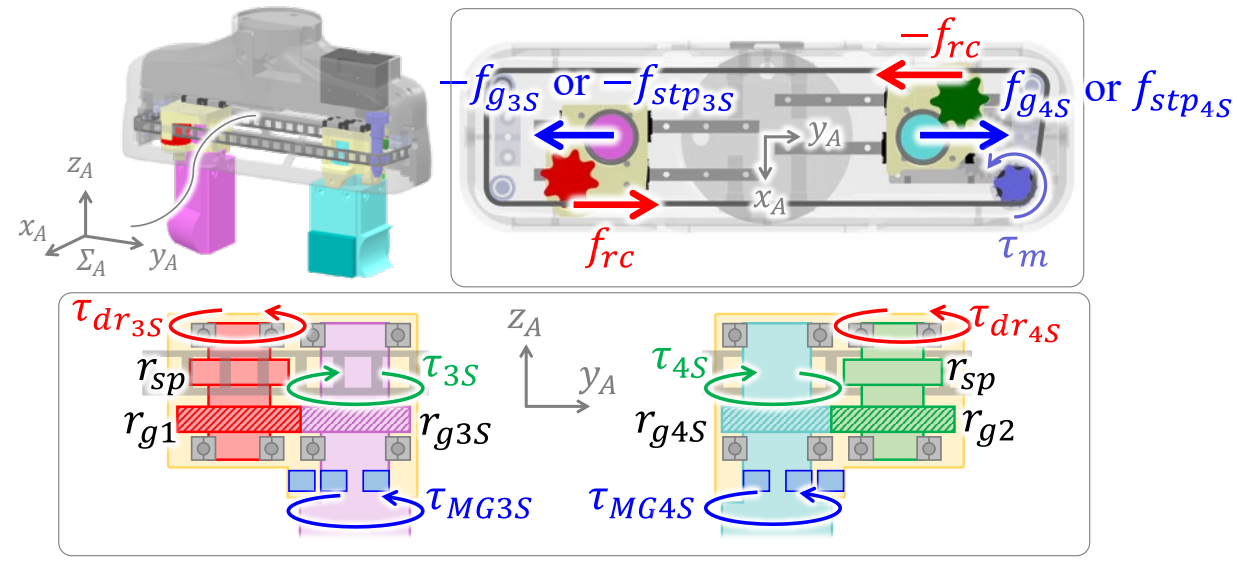

**Fig. 4.** Mechanical relationship of the developed gripper

### *C. Statics*

#### *1) Analysis*

This section details the behavior of the developed gripper with the self-motion-switching mechanism through statics. Fig. 4 shows the mechanical relationships of the gripper. The coordinate frame $\Sigma_A$ is set as shown in Fig. 4. If $\tau_m$ is the input motor torque, the tensional force $f_{rc}$ applied to the roller chain through the input shaft is expressed by:

$$f_{rc} = \tau_m / r_{IS} \tag{1}$$

where $r_{IS}$ denotes the pitch radius of the sprocket part in the input shaft. First, the case of grasping an object is considered.

Let $f_{g_i}$ ($i = \{3S, 4S\}$) be the grasping forces of the 3S and 4S finger units. The force relationship is given by:

$$f_{g_{3S}} = f_{g_{4S}} = f_{rc} = \tau_m / r_{IS} \tag{2}$$

As described in Section II.B, the ratchet mechanism within the finger units prevents the rotation of the finger bodies while grasping the objects. Consequently, any torque applied to the finger bodies is balanced by the torque generated by the ratchet mechanism, or an opening motion is generated.

Next, the case in which the finger units are fully opened and in contact with the stoppers is considered. If letting $f_{stp_i}$ ($i = \{3S, 4S\}$) be the contact forces applied from the stopper to each finger unit, the relationship between the forces is given by

$$f_{stp_{3S}} = f_{stp_{4S}} = f_{rc} = \tau_m / r_{IS} \tag{3}$$

Any force applied in the direction of the opening is balanced using $f_{stp_i}$. When the tensional force $f_{rc}$, is applied to the driving shafts 1 and 2, the torque, $\tau_{dr_i}$ ($i = \{3S, 4S\}$), is also generated at the driving shafts due to $f_{rc}$, expressed as:

$$\tau_{dr_i} = r_{sp} f_{rc} \tag{4}$$

where $r_{sp}$ is the pitch radius of the sprocket part of the driving shaft in both the 3S and 4S finger units, assuming that $r_{sp}$ values for the 3S and 4S finger units are equivalent. Let $r_{g1}$ and $r_{g2}$ be the pitch radii of the gear parts of the driving shafts 1 and 2, respectively, and $r_{g3S}$ and $r_{g4S}$ be the pitch radii of the gear parts of the 3S and 4S finger bodies, respectively. Considering the gear engagement between the driving shaft and finger body in each finger unit, the torques, $\tau_{3S}$ and $\tau_{4S}$, applied to the 3S and 4S finger bodies from the driving shafts are expressed as:

$$\begin{aligned} \tau_{3S} &= \frac{r_{g3S}}{r_{g1}} \tau_{dr_{3S}} = \frac{r_{g3S} r_{sp}}{r_{g1}} f_{rc} = \frac{\alpha_1}{r_{IS}} \tau_m \\ \tau_{4S} &= \frac{r_{g4S}}{r_{g2}} \tau_{dr_{4S}} = \frac{r_{g4S} r_{sp}}{r_{g2}} f_{rc} = \frac{\alpha_2}{r_{IS}} \tau_m \end{aligned} \tag{5}$$

where $\alpha_1 = r_{g3S} r_{sp} / r_{g1}$ and $\alpha_2 = r_{g4S} r_{sp} / r_{g2}$. Let $\tau_{MG3S}$ and $\tau_{MG4S}$ be the torques on the 3S and 4S finger bodies, respectively, generated by the magnetic attraction. The moment balances in the 3S and 4S finger bodies are expressed as:

$$\begin{aligned} \tau_{3S} + \tau_{MG3S} = 0 \to |\tau_{MG3S}| = |\alpha_1 f_{stp_{3S}}| = |\alpha_1 \tau_m / r_{IS}| \\ \tau_{4S} + \tau_{MG4S} = 0 \to |\tau_{MG4S}| = |\alpha_2 f_{stp_{3S}}| = |\alpha_2 \tau_m / r_{IS}| \end{aligned} \tag{6}$$

The larger $\tau_m$ is, the larger $\tau_{MG3S}$ and $\tau_{MG4S}$. It is noteworthy that both $\tau_{MG3S}$ and $\tau_{MG4S}$ have upper limits of $\tau_{MG}^{max}$, as described later. Subsequently, by rotating the motor further in the direction of the finger opening, (6) is lost, leading to finger body rotation and subsequent finger surface or GC switching. The condition is expressed as:

$$\begin{aligned} |\tau_{3S}| = |\alpha_1 f_{stp}| = |\alpha_1 \tau_m / r_{IS}| > |\tau_{MG}^{max}| \\ |\tau_{4S}| = |\alpha_2 f_{stp}| = |\alpha_2 \tau_m / r_{IS}| > |\tau_{MG}^{max}| \end{aligned} \tag{7}$$

Here, the magnetic torques, $\tau_{MG3S}$ and $\tau_{MG4S}$, are analyzed. The mechanical model is shown in Fig. 5. The magnetic attraction between the magnet installed on the finger base and that installed on the finger body is considered. The nominal state is defined as the state in which the distance between the magnets installed in the finger body and base is minimized. The state where the finger body is rotated with $\theta_{FB}$ from the nominal position is considered, as shown in Fig. 5(a). The magnets are assumed to be sufficiently small as point magnetic charges. The analytical model is shown in Fig. 5(b). The base coordinate frame, $\Sigma_B$, for the analysis is set as shown in the figure. Let $P_o$, $P_A$, and $P_B$ be the positions of the rotational center of the finger body, the magnet in the finger base, and the magnet in the finger body, respectively. The magnitude of the magnetic attractive force $f_{MG}$, applied between the magnets is expressed as

$$f_{MG} = \frac{k_m}{\|\boldsymbol{p}_{BA}\|^2} \tag{8}$$

where $k_m$ denotes the constant coefficient determined by the magnetic properties of the magnets and ambience, and $\boldsymbol{p}_{ij}$ is the position vector from $P_i$ to $P_j$ ($i, j \in \{O, A, B\}$). Considering that the magnetic force applied to the magnet in the finger body acts in the direction from $P_B$ to $P_A$, the moment, $\tau_{MG}$, applied to the finger body due to $f_{MG}$ is expressed as:

$$\tau_{MG} = \boldsymbol{p}_{OB} \otimes \frac{f_{MG}}{\|\boldsymbol{p}_{BA}\|} \boldsymbol{p}_{BA} \tag{9}$$

where $\boldsymbol{v}_1 \otimes \boldsymbol{v}_2$ for vectors $v_1 = [v_{1x}, v_{1y}]^{\mathrm{T}}$ and $v_2 = [v_{2x}, v_{2y}]^{\mathrm{T}}$ represents:

$$\boldsymbol{v}_1 \otimes \boldsymbol{v}_2 = v_{1x} v_{2y} - v_{1y} v_{2x} \tag{10}$$

Let $r_{FB}$ be the radius of the cylindrical part of the finger body in which the magnet is installed, and $d_{MG}$ be the gap between $P_B$ and $P_A$ in the nominal state, as shown in Fig. 5(a). $\boldsymbol{p}_{OB}$ and $\boldsymbol{p}_{BA}$ are geometrically given by:

$$\begin{aligned} \boldsymbol{p}_{OB} &= [r_{FB} \cos\theta_d, r_{FB} \sin\theta_d]^{\mathrm{T}} \\ \boldsymbol{p}_{BA} &= [d + r_{FB}(1 - \cos\theta_d), -r_{FB} \sin\theta_d]^{\mathrm{T}} \end{aligned} \tag{11}$$

Then, $\tau_{MG}$ can be derived as follows:

$$\tau_{MG} = -\frac{k_m r_{FB} (r_{FB} + d_{MG}) \sin\theta_{FB}}{(d_{MG}^2 + 2r_{FB}^2 + 2d_{MG} r_{FB} - 2r_{FB}(r_{FB} + d_{MG}) \cos\theta_{FB})^{\frac{3}{2}}} \tag{12}$$

In the prototype design, $r_{FB}$ and $d_{MG}$ were set to 14 and 1 mm, respectively. Considering the properties of the magnet used (TRUSCO, T06R06-M1.6) and the surrounding air environment, $k_m$ was calculated to be $1.07 \times 10^{-5}$ Nmm$^2$. Using these parameters, Fig. 6 shows the derived $|\tau_{MG}|$ from (12). $|\tau_{MG}|$ sharply increases as $\theta_{FB}$ increases from the nominal state until it reaches 2.7°. The magnetic torque reaches its maximum when $\theta_{FB} = 2.7°$ and then decreases sharply with the further increase in $\theta_{FB}$ beyond 2.7°. Hence, $|\tau_{MG}^{max}| = |\tau_{MG}(\theta_{FB} = 2.7°)|$. $|\tau_{3S}|$ and $|\tau_{4S}|$ increase due to the contact between the finger base and the stopper, and if (7) is satisfied, the rotation of the finger body is induced. During this rotation, the magnitude of the rotational angle $\theta_{FB}$ of the finger body is determined by the motor rotational angle $\theta_m$. Let $\theta_{FB3S}$ and $\theta_{FB4S}$ be the $\theta_{FB}$ values of the 3S and 4S fingers, respectively. They are expressed as:

$$\theta_{FB3S} = \frac{r_{g1}}{r_{g3S}} \frac{r_{IS}}{r_{sp}} \theta_m, \theta_{FB4S} = \frac{r_{g2}}{r_{g4S}} \frac{r_{IS}}{r_{sp}} \theta_m \tag{13}$$

The surfaces of both fingers should be simultaneously oriented antipodally. To achieve the antipodal orientation of the finger surfaces, the ratio of $\theta_{FB3S}$ and $\theta_{FB4S}$ should be equal to the ratio of the 360°/3 and 360°/4:

$$\theta_{FB3S} : \theta_{FB4S} = 360°/3 : 360°/4 \tag{14}$$

then,

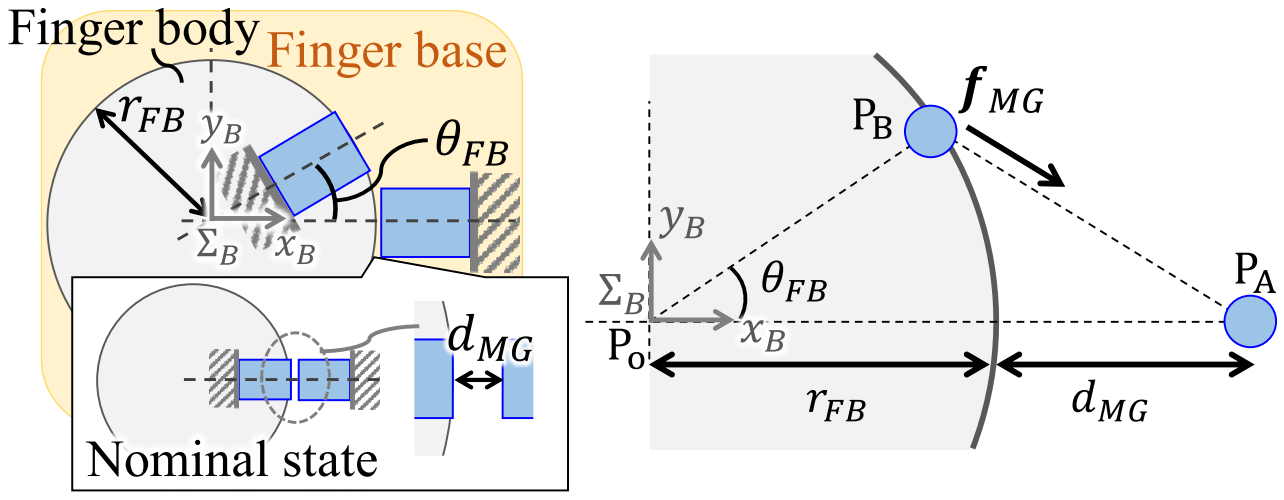

(a) Considered state (b) Simplified model

**Fig. 5.** Mechanical model for analyzing the magnetic torque

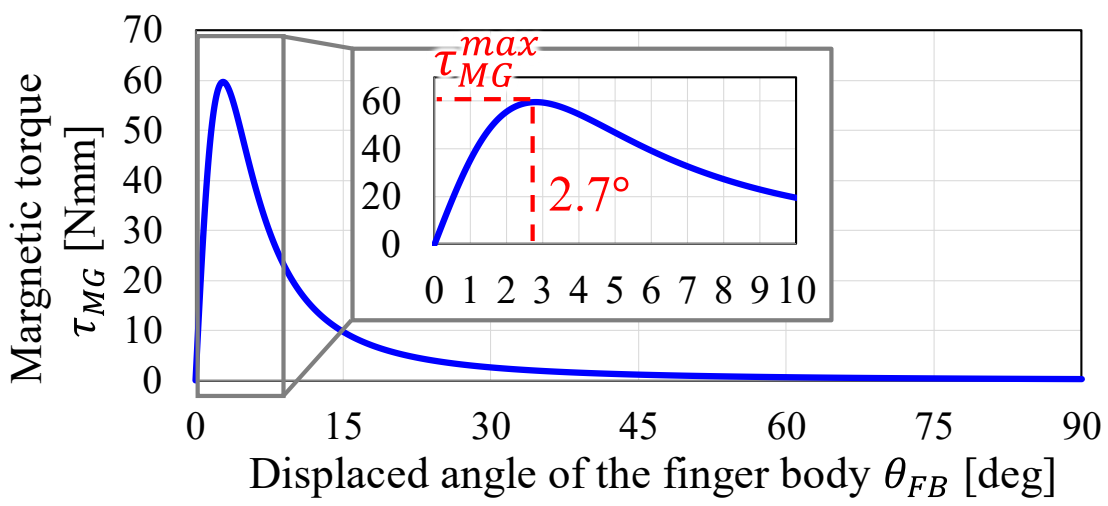

**Fig. 6.** Analytical magnetic torque $\tau_{MG}$ derived from (12)

$$\frac{r_{g1}}{r_{g3S}} : \frac{r_{g2}}{r_{g4S}} = \frac{1}{3} : \frac{1}{4} \tag{15}$$

Configuring the gear ratios to satisfy (15) allows for the simultaneous antipodal orientation of the two finger surfaces with every rotation of the motor by an angular interval $\Delta\theta_{sw}$, as given by:

$$\frac{360°}{3} = \frac{r_{g1}}{r_{g3S}} \frac{r_{IS}}{r_{sp}} \Delta\theta_{sw} \text{ or } \frac{360°}{4} = \frac{r_{g2}}{r_{g4S}} \frac{r_{IS}}{r_{sp}} \Delta\theta_{sw} \tag{16}$$

The design parameters were set as follow: $r_{IS}$=20 mm, $r_{sp}$=15 mm, $r_{g1}$=10 mm, $r_{g3S}$=12 mm, $r_{g2}$=7.5 mm, and $r_{g4S}$=12 mm, then $\Delta\theta_{sw}$ is derived as 108°.

In the prototype design, the finger surfaces were set such that the surface was switched in the order of flat, convex, and concave surfaces in the 3S finger and flat, convex, concave, and deformable surfaces in the 4S finger. The GC mode, in which the surfaces of both the 3S and 4S fingers were flat, was defined as the initial mode. The transition of the GC mode according to the motor rotation is shown in the attached video. All combinations of finger surface shapes are realized through the repeated rotations of the finger. Considering that the two fingers are rotated at different angles by a single-motor drive, the number of available combinations, i.e., the total number of available GC modes ($n_{GC}$), is determined by

$$n_{GC} = \begin{cases} n_a , & n_a \bmod n_b = 0 \\ \mathrm{LCM}(n_a, n_b), & otherwise \end{cases} \tag{17}$$

where $n_a$ and $n_b$ denote the numbers of surfaces on the two fingers with $n_a \geq n_b$, and $\mathrm{LCM}(n_a, n_b)$ denotes the least common multiple of $n_a$ and $n_b$. As described in Section II.B, to maximize $N_{GC}$ while decreasing $n_a$ and $n_b$, the numbers of the surfaces were set to four and three.

*2) Validation*

The proposed mechanism has been constructed based on the analysis presented in the previous section. This section presents the experimental validation conducted to verify that the mechanism exhibits the desired theoretical behavior. To evaluate the motion behavior of the gripper, two experiments were conducted, focusing on the grasping and GC mode switching operations. First, the experiment for the grasping operation is described. The experimental setup is illustrated in Fig. 7. The setup represents the configuration of the driving mechanism of the 3S finger unit. The marker was mounted on the top side of the finger body, and the camera was fixed at the top of the setup. The moving distance of the finger unit, denoted as $d_f$, and the rotational angle of the finger body, $\theta_{FB}$, were derived through the image processing of the position and orientation of the marker captured by the camera. A force gauge was fixed to the part corresponding to the gripper body to measure the grasping force $f_{g_{3S}}$, during the finger-closing operation. The finger unit was initially positioned 10 mm from the indenter of the force gauge. In the experiment, the motor rotated in the direction of closing the finger until the motor torque reached 800 Nmm. The motor torque, $\tau_m$, was also monitored during the experiment. The experiment was repeated 10 times. The results are presented in Fig. 8. The analytical $f_{g_{3S}}$ (dashed line) derived from (2) using measured $\tau_m$ is also shown. As the motor rotated, the finger unit firstly translated. If the finger translation was terminated due to contact between the finger and force gauge, $f_{g_{3S}}$ began to increase. $f_{g_{3S}}$ increases with an increase in $\tau_m$. While the finger unit translated and was in contact with the force gauge, the ratchet mechanism prevented any rotation of the finger body. The analytical $f_{g_{3S}}$ was close to the measured value. Next, the experiment on the GC mode switching is described. The experimental setup was the same as that shown in Fig. 7 except that the force gauge was not used. Alternatively, the motor angle, $\theta_m$, was measured during the experiment, in addition to $\tau_m$, $d_f$, and $\theta_{FB}$. The finger unit was initially positioned 3 mm away from the stopper. The motor rotated in the direction to open the finger until the monitored $\theta_{FB}$ reaches 120° that is the angle required for the finger surface to transition to the subsequent finger surface in 3S finger. The results are presented in Fig. 9. The finger unit was translated until the finger base contacted the stopper, without rotating the finger bodies. The results obtained thus far, together with those shown in Fig. 8, suggest that there was no undesired rotation of the finger bodies during their opening and closing actions. Subsequently, the finger body began to rotate with further motor rotation. The angular displacement, $\Delta\theta_{sw}$, of the motor rotation required for the finger surface transition to the subsequent one ($\theta_{FB} \rightarrow 120°$) was 108.5°, almost identical to the theoretical value (i.e., 108°) shown in (16). At the moment when the rotation of the finger body occurred, $\tau_m$ reached the peak value. The torque applied from the drive shaft to the finger body ($\tau_{3S}$) was derived from (5), using the measured $\tau_m$. The $\tau_{3S}$ and analytical $\tau_{MG}$ calculated in Fig. 6 are shown in Fig. 10. $\tau_{3S}$ reached its peak value at $\theta_m = 3.7°$ that is close to the angle at which the $\tau_{MG}$ reaches its maximum. The gradual decrease in $\tau_{3S}$ compared to that in $\tau_{MG}$ is attributed to friction within the driving mechanism, such as the roller chain, disturbing the rotational motion of the finger body. Nevertheless, these results validate the analysis presented in the previous section and confirm the desired behavior of the proposed mechanism.

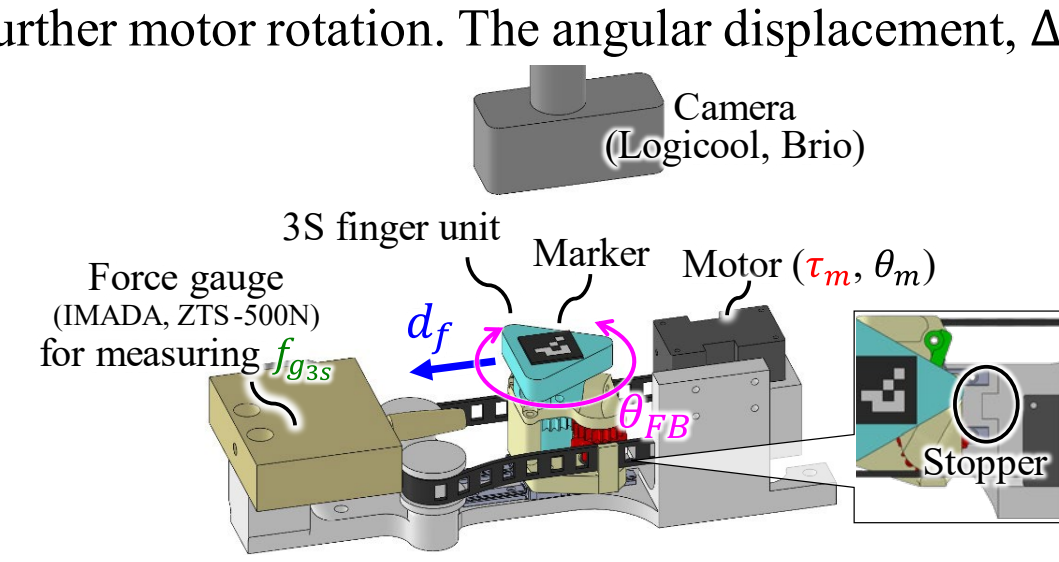

**Fig. 7.** Experimental setup for evaluating the proposed mechanism

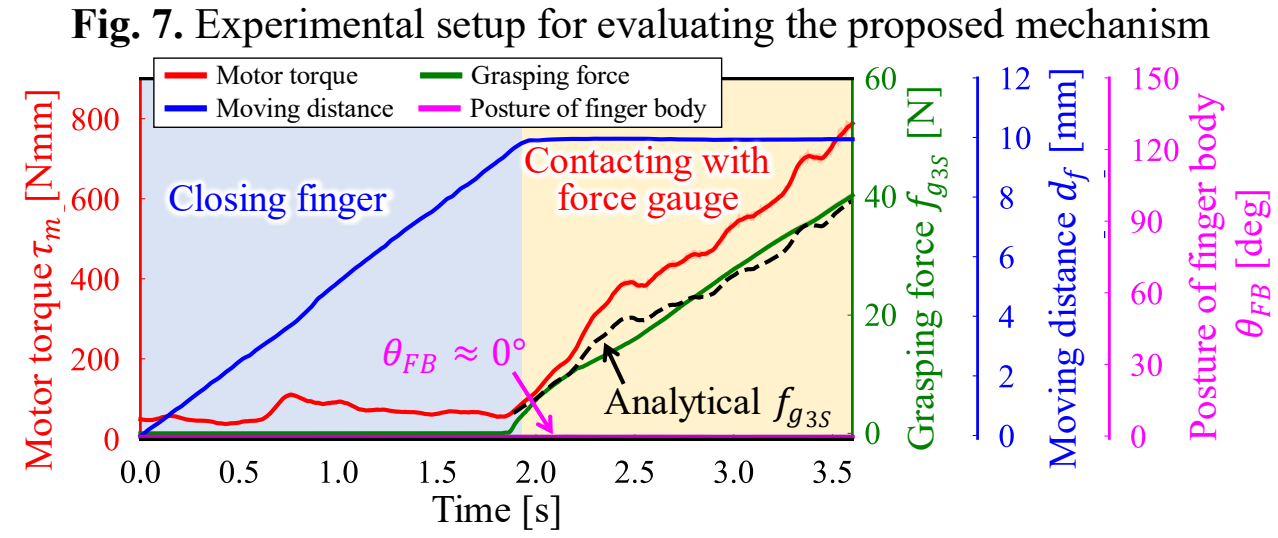

**Fig. 8.** Result of the evaluation for the grasping operation

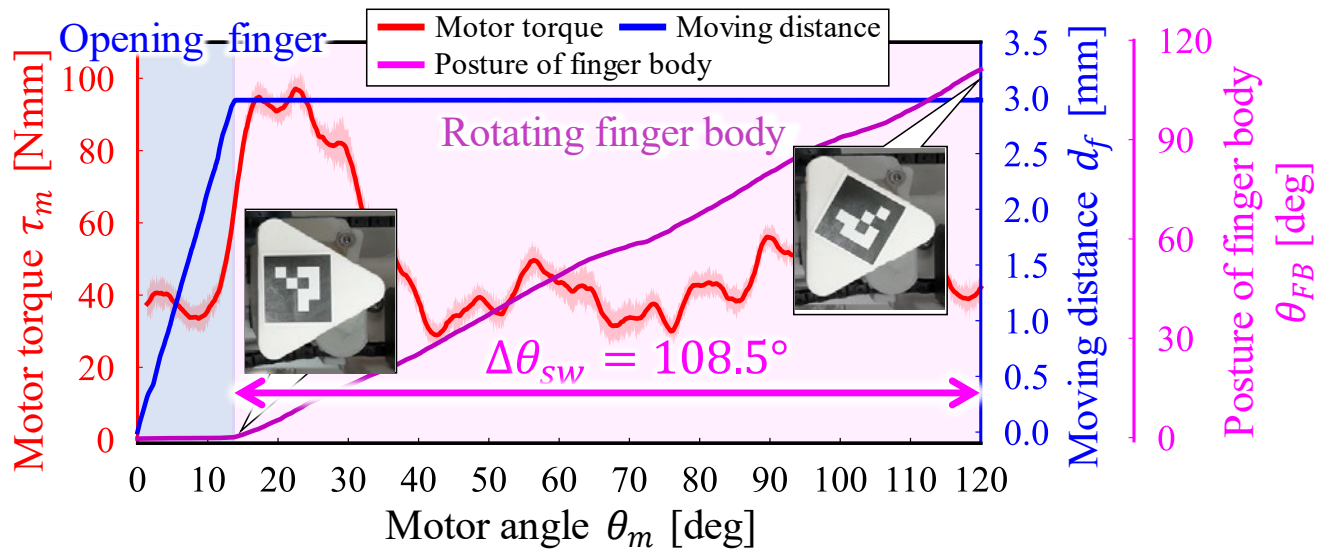

**Fig. 9.** Result of the evaluation for the GC mode switching operation

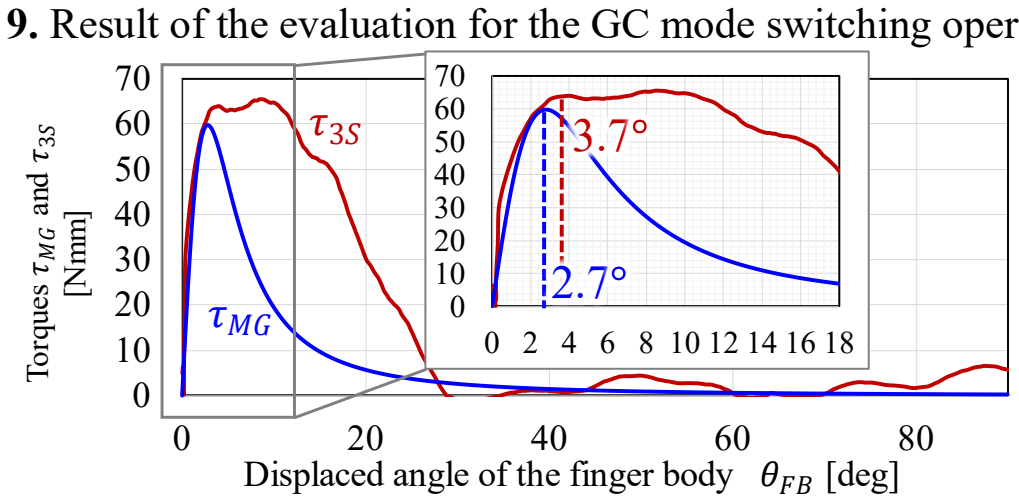

**Fig. 10.** Comparison of analytical $\tau_{MG}$ and measured $\tau_{3S}$ during finger body rotation

### *D. Design of finger surface shapes*

This section describes the strategy for designing the finger surface shapes. The grasping motion of the parallel-jaw grippers, including the developed gripper, is restricted to the linear opening and closing of the fingers. Thus, for the 3S and 4S fingers, considerably different finger surfaces were adopted to enhance versatility. The radii of the convex and concave surfaces were set to be identical, ensuring that the surfaces fit each other and enhancing the stable grasping when the concave and convex surfaces were employed on each finger. The remaining surface of the 4S finger was set as a deformable flat surface to ensure a high adaptability to complex object shapes. Note that the deformable surface could cause uncertainties in the position and posture of the grasped object; thus, the other three rigid surfaces are prioritized for use. The radii of the concave and convex surfaces were determined based on the height of the target object. The height of the target objects was mainly less than 20 mm, and then $r_f$ was set to 10 mm.

## III. Control Methodology

This section describes the control methodology used to realize the grasping motion and the desired GC mode. The proposed control methodology is shown in Fig. 11. In this methodology, the motor torque control is used to close the fingers to grasp an object. The motor position control is used to open the fingers for releasing the object and to switch the GC mode. In the initial state, the fingers are fully opened; i.e., the finger bases are in contact with the stopper on the gripper body. $\theta_m^{ini}$ is the motor position ($\theta_m$) in the initial state. To grasp an object with the target grasping force $f_g^{goal}$, the target motor torque $\tau_m^{goal}$ to achieve the $f_g^{goal}$, is derived using (2). By rotating the motor in the forward direction such that the motor torque reaches $\tau_m^{goal}$, the finger is closed and the object is grasped with the desired grasping force $f_g^{goal}$. To switch the GC mode from the present one to the target one, the target angle of the motor rotation, $\theta_m^{goal}$ is derived. The present GC mode number is denoted as $k_{now}$, and the target GC mode number is denoted as $k_{goal}$. $\theta_m^{goal}$ to realize the GC mode with number of $k_{goal}$ is given by

$$\theta_m^{goal} = \begin{cases} (k_{goal} - k_{now})\Delta\theta_{sw} & if\ k_{goal} \geq k_{now} \\ (n_{GC} + k_{goal} - k_{now})\Delta\theta_{sw} & otherwise \end{cases} \quad (18)$$

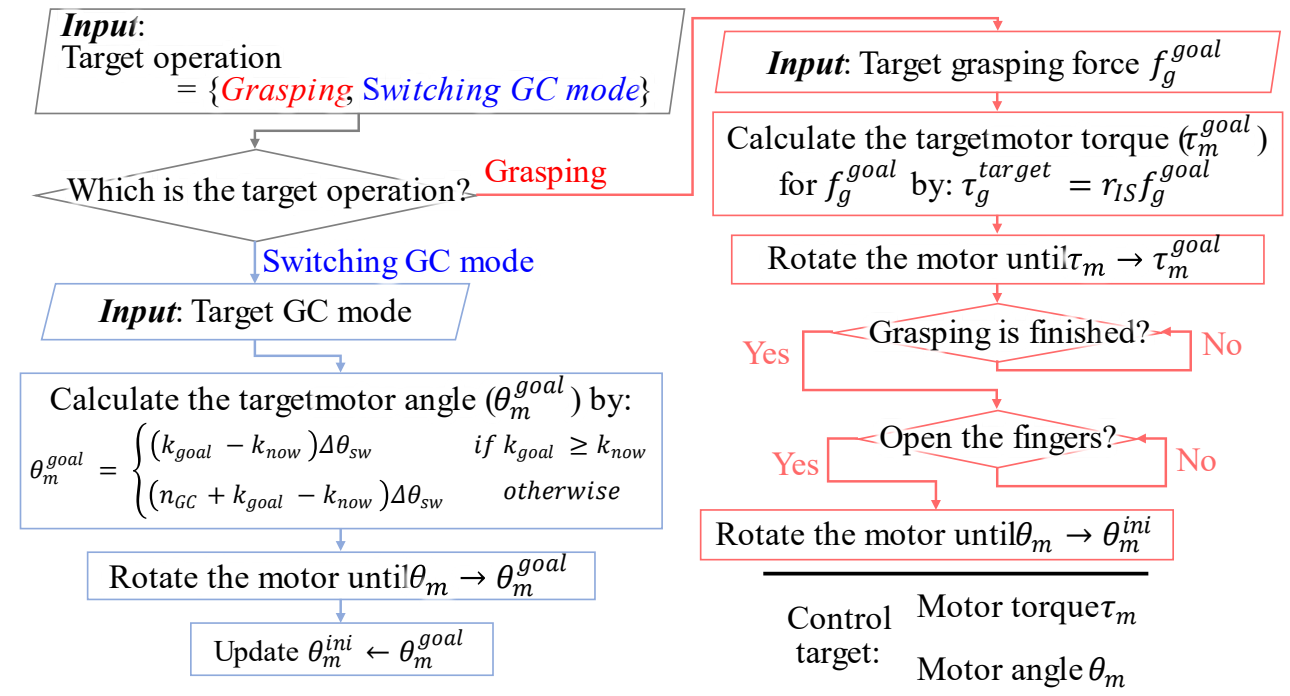

**Fig. 11.** Control methodology for grasping and GC mode switching operations

As a reminder, $n_{GC}$ denotes the total number of the available GC modes. The target GC mode is achieved by controlling the motor position to the $\theta_m^{goal}$ given by (18). $\theta_m^{ini}$ is updated for the subsequent operations: $\theta_m^{ini} \leftarrow \theta_m^{goal}$.

## IV. Experimental Evaluation

This section introduces the strategy for selecting the GC mode and presents the grasping tests. First, the fundamental grasping test was conducted for primitive-shaped objects. The GC mode selection strategy is proposed based on the grasping test results. This strategy was then applied to subsequent grasping tests with industrial objects to evaluate the performance of the developed gripper.

### *A. Grasping test of primitive shaped objects*

To evaluate the grasping ability of the developed gripper in each GC mode, the grasping tests were conducted on several primitive-shaped objects, as shown in Fig. 12. The gripper was attached to an automatic positioning stage. The target object was placed on a table at several different postures. The positioning stage was moved down to a position where the finger was in slight contact with the table, and the gripper grasped the object using the control methodology presented in Section III. The target grasping force was set to 20 N. After grasping, the stage was moved upward, and the results (whether grasping was successful) were observed. Successful grasping is defined as the lifting and holding of an object (rotation after lifting is not considered). The representative results are presented herein. First, the results for the fundamental grasping configuration, i.e., the GC modes of GC1-3 defined in Fig. 1(b), are explained (Fig. 13). In the case of GC1, all the objects were successfully grasped. This GC mode is effective for grasping small and thin objects. The observed grasping style was close to that of precision grasping [20]. In this case, force-closure grasping was performed, and the grasping stability depended on the frictional condition of the surface. When both finger surfaces were set to convex or concave shapes, i.e., GC2 or GC3, grasping was achieved through geometric constraints. If the grasped object is immobilized, the grasping is of form closure. Otherwise, it is caging. The self-alignment was observed when the form closure was constructed. The self-alignment is effective for improving the efficiency of assembly tasks [21]. When grasping the large cylinder (No. 1) with GC3, the grasping was accomplished with four contact points, while the object position was aligned. In contrast, when grasping the small cylinder (No. 2) with GC3, the object was so small that its position could not be restricted by the contact with the finger surfaces. Nevertheless, the caging was achieved by confining the movable range of the object within the area enclosed by the fingers. The geometrical constraints in the grasping provide a large payload, regardless of the frictional condition. When the thin plate (No. 4) was grasped with GC3, two states of the object posture were observed after grasping. First, the object was pinched by the tip of the finger surface. Second, the object was tilted during the grasping operation. This difference would be caused by the positioning error of the automatic positioning stage. This uncertainty in the posture of the grasped object is undesirable for subsequent operations such as placing. In summary, convex and concave finger surfaces are effective for firmly grasping an object using the geometrical constraints between the finger surfaces and the object. These surfaces are particularly advantageous in situations that demand a substantial resistible force, such as when handling heavy objects, as opposed to flat fingers whose grasping stability relies on frictional conditions. However, the convex and concave finger surfaces are not preferred for grasping small objects. The grasping test with the GC modes, in which the two finger surfaces are different from each other, demonstrated that these GC modes are effective for grasping asymmetrically shaped objects and grasping multiple objects together (Fig. 14(a)). In the grasping test employing GC4 with the deformable surface, an inability to grasp a thin plate was observed. The deformable surface was deformed due to contact with the object, resulting in slippage between the finger and the object (Fig. 14(b)). This result indicates that the deformable surface is not preferable for grasping small and thin objects.

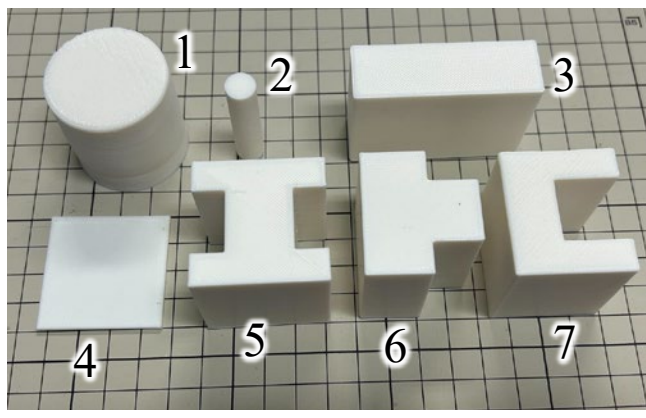

| No. | Character | Size |
|---|---|---|
| 1 | Large cylinder | Dia. 40 mm |
| 2 | Small cylinder | Dia. 10 mm |
| 3 | Rectangular | W60×H20 mm |
| 4 | Thin plate | Thickness 1 mm |
| 5 | H-shaped object | W40×H40 mm |
| 6 | T-shaped object | W40×H35 mm |
| 7 | U-shaped object | W40×H35 mm |

**Fig. 12.** Target objects for the grasping test of primitive shaped objects

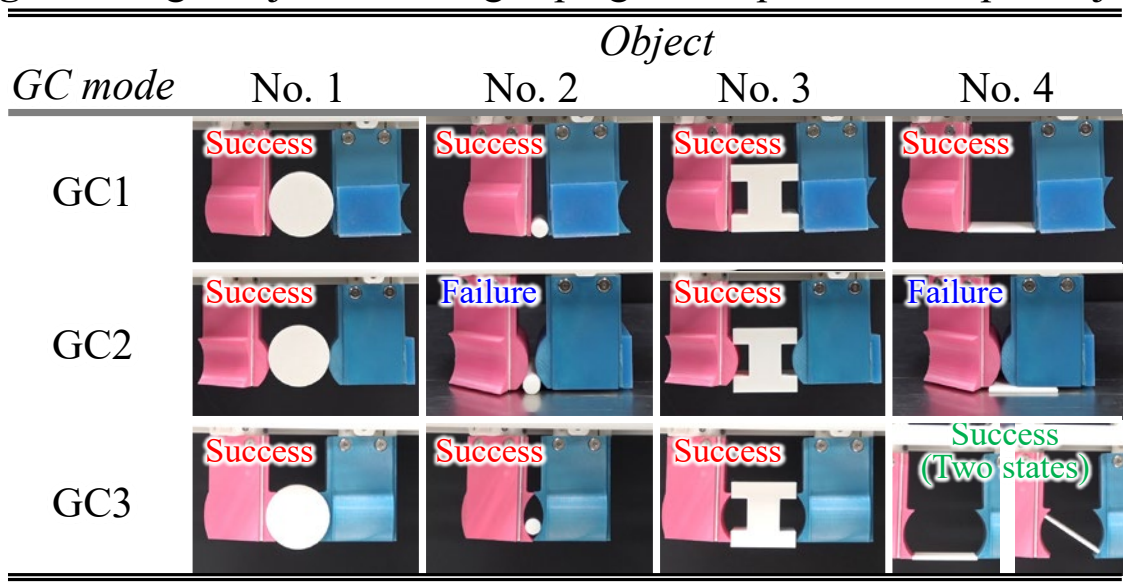

**Fig. 13.** Grasping test results with the GC1, GC2, and GC3

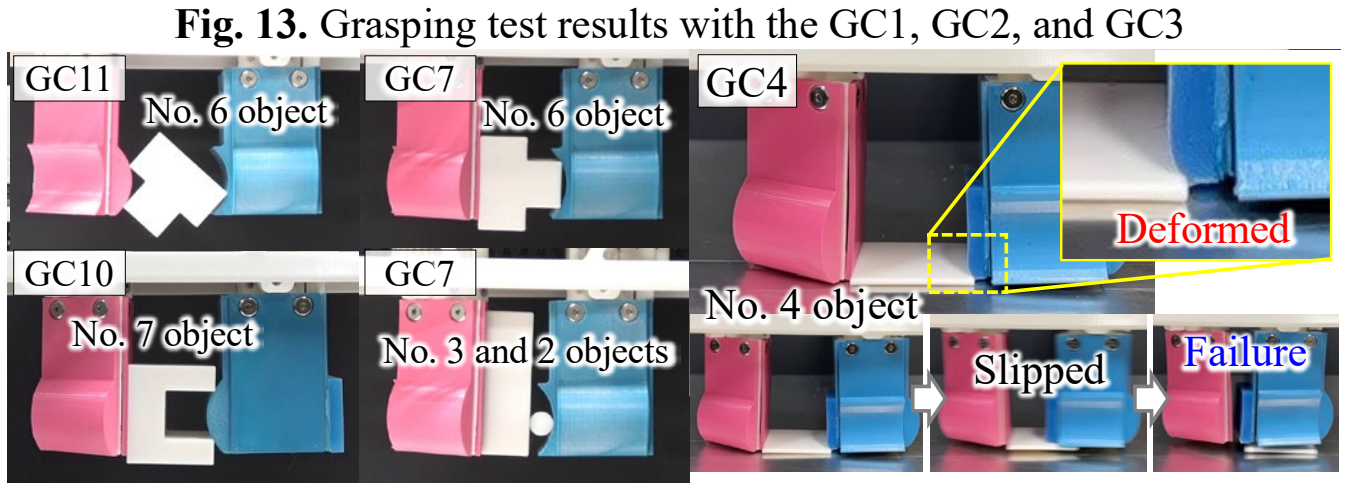

(a) GC modes with different surfaces (b) GC mode with deformable sutface

**Fig. 14.** Grasping test result with the GC mode with different surfaces

### *B. GC Mode Selection Strategy*

Based on the grasping test described in the previous section, a strategy for selecting the GC mode (combination of finger surfaces) is presented. A strategy based on the shape of the target object was adopted to select the finger surfaces. The object shape, including the thickness and shape/contour of the area facing each finger, is assumed to be given. Initially, the proposed strategy narrows down the candidate finger surface shapes using the two criteria shown in Fig. 15: 1) the candidate is chosen according to each surface of the target object facing the finger, and 2) the candidate is selected based on the thickness and height of the target object. From the grasping test results in the previous section, the flat finger surface can be adopted to grasp objects of various shapes, whereas convex and concave finger surfaces are preferable for the concave and convex surfaces of the object, respectively. It is difficult to use the convex finger surface to grasp a small object (see Criterion 2). If two or more GC modes satisfy these criteria, the optimal GC mode is determined as the one among them achieved with the smallest motor rotation angle from the current GC mode. This selection policy minimizes the time required to switch the GC mode, thereby enhancing productivity. If no GC mode satisfies either condition, a deformable surface is selected to increase the possibility of successful grasping.

### *C. Grasping test of industrial objects*

This section presents the grasping tests for various objects, including industrial items. The experimental setup was the same as that described in Section IV.A. The GC mode was determined using the proposed GC mode selection strategy. The results are shown in Fig. 1(b). All objects, including a heavy toolbox (5 kg), thin nut (thickness: 1 mm), and asymmetrical objects (motor with pulley and bearing holder), were lifted and held successfully . The results validated the effectiveness of the developed gripper for grasping various objects through the GC mode switching. The activation of the GC mode switching limits the grasping force when grasping an object using the backside of the fingers; however, grasping with the backside of the fingers remains feasible. Lifting a lightweight box was achievable under this condition, as shown in the attached video.

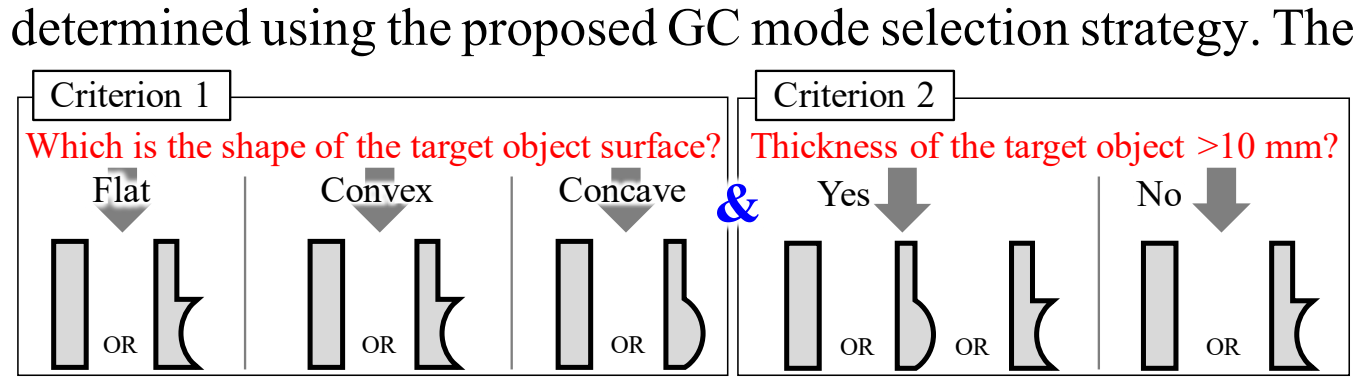

**Fig. 15.** Criteria for selecting the finger surface shape

## V. Discussion

In this study, the finger surfaces were designed by focusing on 2D planar grasping ($yz$-plane in Fig. 1(a)) to facilitate the understanding of the design concept. Additional functions could be achieved by using different finger surfaces. For example, the $xy$-planar alignment of the position of the object can be achieved by adopting the finger surface shown in Fig. 16 (a). In the GC mode-selection strategy described in Section IV, the selection of the desired finger surfaces is influenced only by the shape of the target object. However, other factors, such as the content of the target task, can be incorporated into the selection strategy. As described in the grasping test section, three types of grasping were achieved: force-closure, form-closure, and caging. The grasping-type selection based on the target task can be integrated into the finger surface selection process and incorporated into the criteria shown in Fig. 15. Under this setting, a more suitable GC mode could be selected according to the target object and task. This study focused primarily on objects with simple shapes. The GC-mode selection based on sensor information would be helpful for stably grasping objects with more complex shapes. Fig. 16(b) shows the finger design integrated with a time-of-flight-based proximity sensor to measure the object shape near the finger. Using this sensor information and grasping quality evaluation method [22], the GC mode could be determined for objects with more complex shapes. Finally, an experiment was conducted to perform the GC mode switching 100,000 times. The proposed mechanism worked without significant damage after the experiment, thereby demonstrating the high durability of the gripper owing to the magnet-based switching mechanism.

## VI. Conclusion

This study proposed a novel robotic gripper designed to facilitate the grasping of various objects by changing the GC mode. The GC modes were changed through the rotation of the fingers with the multiple different surfaces. Both the finger rotation for the GC mode switching and the finger translation for the grasping operation were realized by a single motor, owing to a novel self-motion switching mechanism. If the motor rotates in the direction of the finger opening beyond the fully opened state, the motion shifts from the finger translation to the finger rotation. Magnetic attraction is employed to regulate the force required for this motion switching, ensuring stable switching and finger movement. The incorporation of magnets is also effective in preventing damage such as wear, thus enhancing the overall durability of the gripper. Its durability was experimentally confirmed. The control and GC mode selection methodologies were presented for the developed gripper. The effectiveness of the developed gripper was demonstrated through the grasping tests. Our future works would involve the development of a GC mode-selection strategy using sensors.

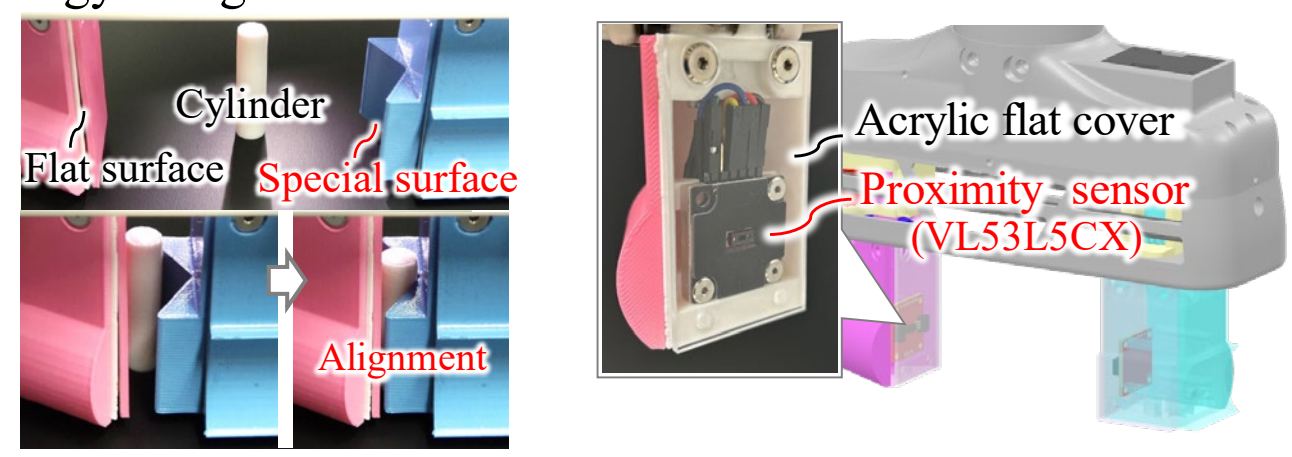

(a) Surface for $xy$-planar-alignment (b) Gripper with proximity sensor

**Fig. 16.** Ideas for future works